\documentclass{article}

\usepackage[dblblindworkshop,final]{neurips_2025}

\usepackage[utf8]{inputenc} 
\usepackage[T1]{fontenc}    
\usepackage{hyperref}       
\usepackage{url}            
\usepackage{booktabs}       
\usepackage{amsfonts}       
\usepackage{nicefrac}       
\usepackage{microtype}      
\usepackage{xcolor}         
\usepackage{graphicx}
\usepackage{tabularx}
\usepackage{xcolor}
\usepackage{xspace}
\usepackage{amsmath}

\newcommand{\algname}{\textsc{Wander}\xspace}
\newcommand{\algnamene}{\textsc{Wander-NE}\xspace}
\newcommand{\algnamefe}{\textsc{Wander-FE}\xspace}

\title{Evolve to Inspire: Novelty Search for Diverse Image Generation}

\workshoptitle{The First Workshop on Generative and Protective AI for Content Creation}

\author{
  Alex Inch \\
  University College London\thanks{Now at the University of Oxford} \\
  \texttt{alex.inch@eng.ox.ac.uk}
  \And 
  Passawis Chaiyapattanaporn \\
  University College London \\ 
  \texttt{official.passawis@gmail.com}
  \And
  Yuchen Zhu \\
  University College London\thanks{Now at Tesco Technology} \\ 
  \texttt{yuchen.zhu.24@ucl.ac.uk}
  \And
  Yuan Lu \\
  University College London\thanks{Now at Microsoft Research} \\ 
  \texttt{yuan.lu.20@ucl.ac.uk}
  \And
  Ting-Wen Ko \\
  University College London \\ 
  \texttt{ting-wen.ko.24@ucl.ac.uk}
  \And
  Davide Paglieri \\
  University College London \\ 
  \texttt{davide.paglieri.22@ucl.ac.uk}
}

\begin{document}

\maketitle

\begin{abstract}


Text-to-image diffusion models, while proficient at generating high-fidelity images, often suffer from limited output diversity, hindering their application in exploratory and ideation tasks. Existing prompt optimization techniques typically target aesthetic fitness or are ill-suited to the creative visual domain. To address this shortcoming, we introduce \algname, a novelty search-based approach to generating diverse sets of images from a single input prompt. \algname operates directly on natural language prompts, employing a Large Language Model (LLM) for semantic evolution of diverse sets of images, and using CLIP embeddings to quantify novelty. We additionally apply emitters to guide the search into distinct regions of the prompt space, and demonstrate that they boost the diversity of the generated images. Empirical evaluations using FLUX-DEV for generation and GPT-4o-mini for mutation demonstrate that \algname significantly outperforms existing evolutionary prompt optimization baselines in diversity metrics. Ablation studies confirm the efficacy of emitters.

\end{abstract}

\section{Introduction}



Text-to-image diffusion models like Stable Diffusion, FLUX and GLIDE excel at generating visually appealing images from text prompts \cite{rombach_diffusion, flux2024, nichol2022glidephotorealisticimagegeneration}. However, a significant limitation of these models is that it can be difficiult to use them to generate diverse sets of images \citep{marwood2023diversitydiffusionobservationssynthetic} unless specifically directed by a user actively writing specific, diverse prompts. This lack of diversity hinders their utility in applications like ideation or exploration, where quickly generating novel ideas is crucial. Simply repeating the prompt yields similar results, and manually tweaking prompts can lead to unpredictable changes, making systematic exploration difficult. 

Large Language Models (LLMs) have shown promise in generating diverse prompts for text-based tasks via mutation \citep{bradley2023qualitydiversityaifeedback, samvelyan2024rainbowteamingopenendedgeneration, faldor2025omniepicopenendednessmodelshuman}, but their application to image generation has primarily focused on optimizing prompts for certain fitness objective such as aesthetic or Natural Language Processing (NLP) task performance \citep{lm_crossover, hao2023optimizing, wu2024universalpromptoptimizersafe, promptify, chen2024tailoredvisionsenhancingtexttoimage}. This contrasts with open-ended exploration, which prioritizes novelty and exploration instead of convergence to a single `best' image, motivating the need for a dedicated approach to systematically enhance image diversity.

\begin{figure}[h!]
    \centering

    \includegraphics[width=1.0\linewidth]{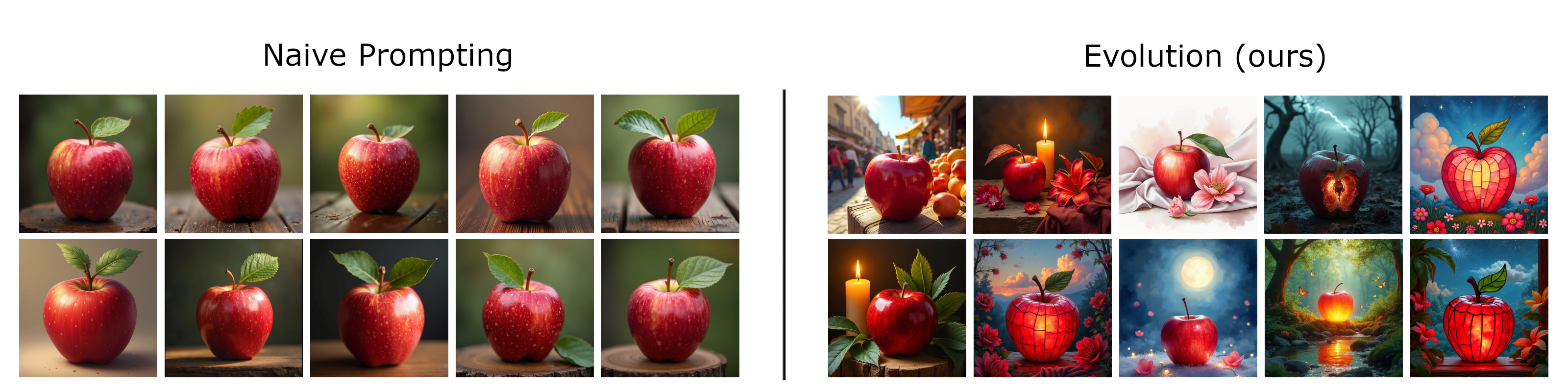}

    \caption{Our method generates significantly more diverse images than reusing a prompt multiple times.}
    \label{fig:qualitative-comparison}
\end{figure}

One prominent diversity-seeking method is Quality Diversity through AI Feedback (QDAIF) \citep{bradley2023qualitydiversityaifeedback}. QDAIF uses an LLM to rate text and assign it to cells on a MAP-Elites grid \citep{mouret2015illuminatingsearchspacesmapping}, evolving it using an LLM to produce a set of diverse texts while maintaining quality. However, adapting these approaches to images is difficult. Our preliminary experiments (Appendix \ref{appendix:attempted}) show that using Vision-Language Models (VLMs) within a QDAIF framework fails; VLM fail to consistently identify qualitatively novel images or accurately categorize images within a MAP-Elites grid based on visual characteristics.



Instead, we propose a novelty search-based approach designed to generate diverse image sets from a single starting point \citep{6793380}. We quantify image novelty using the cosine distance between CLIP image embeddings \citep{radford2021learningtransferablevisualmodels}. As a mutation operator to generate new individuals, we use an LLM.

Inspired by \citep{Fontaine_2020}, we additionally introduce \textit{emitters} to this problem setting. Emitters are specialized mutation strategies that guide evolution into distinct areas of the behavior space. In our case, emitters are prompts that instruct the LLM to mutate a text prompt in specific manners (a full list can be found in appendix \ref{appendix:emitters}).



Our experimental methodology involved comparing \algname against other methods by running each algorithm 10 times on identical prompts. To assess the impact of our design choices, we performed an ablation study over different emitter strategies, conducting 10 runs for each of the 10 initial prompts. Additionally, we evaluated the long-horizon performance of random versus bandit-driven selection by running both methods for 30 generations. Results show that \algname achieves superior image diversity while maintaining reasonable relevance and token efficiency compared to existing prompt optimization baselines. In all, we demonstrate that:
\begin{itemize}
    \item Novelty search using image CLIP embeddings is capable of generating highly-diverse sets of images given a simple text prompt as a starting point.
    \item Introducing human-designed mutation strategies (emitters) enhances the diversity of generated images.
    \item We introduce \algname, a framework for prompt optimization that leverages an LLM as a mutation engine, where randomly sampling from a predefined set of semantic instructions (emitters) proves to be a powerful mechanism for diverse exploration.
\end{itemize}

\section{Related Work}


\subsection{Fine-Tuning for Diversity}
Recent works have investigated fine-tuning language models for more diverse generation, using reinforcement learning or distillation approaches \citep{hao2023optimizing, Zhang2024ForcingDD, Cideron2024DiversityRewardedCD, Miao_2024_CVPR}.

However, fine-tuning approaches have significant limitations; they require access to model parameters, can have high up-front costs, and need to be re-run from scratch for different models. For this work, therefore, we focus on more flexible black-box approaches that can be evaluated via APIs.

\subsection{Evolutionary Strategies}
The use of evolutionary strategies (ES) has a long history in black-box optimization, where iterative mutation and selection mechanisms drive exploration over complex, high-dimensional spaces.

A critical advancement in the evolution literature was the development of novelty search. Unlike traditional objective-driven optimization, novelty search explicitly rewards behavioral diversity, thereby avoiding premature convergence to local optima. The seminal work by \citet{6793380} demonstrated that novelty-guided evolution can outperform objective-based search by encouraging exploration away from deceptive regions of the search space. This was later extended \citep{10.1145/2001576.2001606} to introduce local competition to balance exploration (novelty) with exploitation (performance).

\subsection{Discrete Prompt Optimization}

Another line of work explores discrete prompt optimization. Methods like APE \citep{zhou2022large} adopt a Monte Carlo sampling strategy to iteratively sample and optimize the prompts. APE focuses on exploration while APO focuses more on exploitation by computing gradient with regard to a given sample. Recent advancements better explore the discrete prompt space by incorporating evolutionary strategies, such as genetic differential algorithms and multi-phase mutation \citep{guo2024connectinglargelanguagemodels, cui2024phaseevounifiedincontextprompt}. However, existing methods for discrete prompt optimization focus on maximizing a fitness function, such as quality or NLP task performance on a development set, as opposed to our problem setting of diverse image generation.

\subsection{Open-Ended Prompt Evolution}

Prior work on evolutionary prompt optimization with LLMs has laid a strong foundation for evolving high-quality prompts. \textsc{PromptBreeder} and \textsc{Rainbow Teaming} \citep{promptbreeder, samvelyan2024rainbowteamingopenendedgeneration} use self-referential evolution, in which an LLM simultaneously mutates a population of task- and mutation-prompts. 
These methods have been shown to find novel solutions to problems including arithmetic, common-sense reasoning and jailbreaking LLMs. Unfortunately, owing to a lack of open-source implementations we were unable to evaluate these methods for our use case.

 Additionally, for creative purposes, QDAIF \citep{bradley2023qualitydiversityaifeedback} uses a MAP-Elites-based approach to generate diverse, high-quality text. In this work we assess an extension of QDAIF to image generation using a vision-language model (VLM) for image feedback, but find it is poorly-suited to the task (see Appendix \ref{appendix:attempted}.)

 \subsection{Prompt Rewriting}

 A related problem is prompt-rewriting or caption-upsampling; training ancillary models which rewrite prompts for more aesthetic or diverse image generation \citep{Betker2023BetterCaptions, hao2023optimizing}. The closest work to ours is \citet{datta-etal-2024-prompt}, which trains a ``prompt expander" model to rewrite user prompts. This is a powerful method, but requires compute-heavy dataset generation and training stages. Additionally, since dataset creation requires access to a given image model, prompt expanders must be retrained for each new image model.

\subsection{Concurrent Work}

During the development of this report, concurrent work was released addressing diverse image generation using a similar approach. Lluminate \citep{joel_creative_2025} also uses novelty search to generate diverse results over images. However, rather than task-specific emitters, Lluminate employs generic "creative strategies" drawn from sources such as Oblique Strategies. Lluminate also uses an unbounded genetic pool, while \textsc{Wander} uses a fixed-size pool, which leads to significantly more token-efficient evolution.

In a benchmark comparison, we find that \textsc{Wander} achieves more diverse image pools (as measured by Vendi Score) than Lluminate, while using seven times fewer tokens.

\section{The \textsc{Wander} Framework}
\begin{figure*}[t!]
  \centering
  \includegraphics[width=0.9\textwidth]{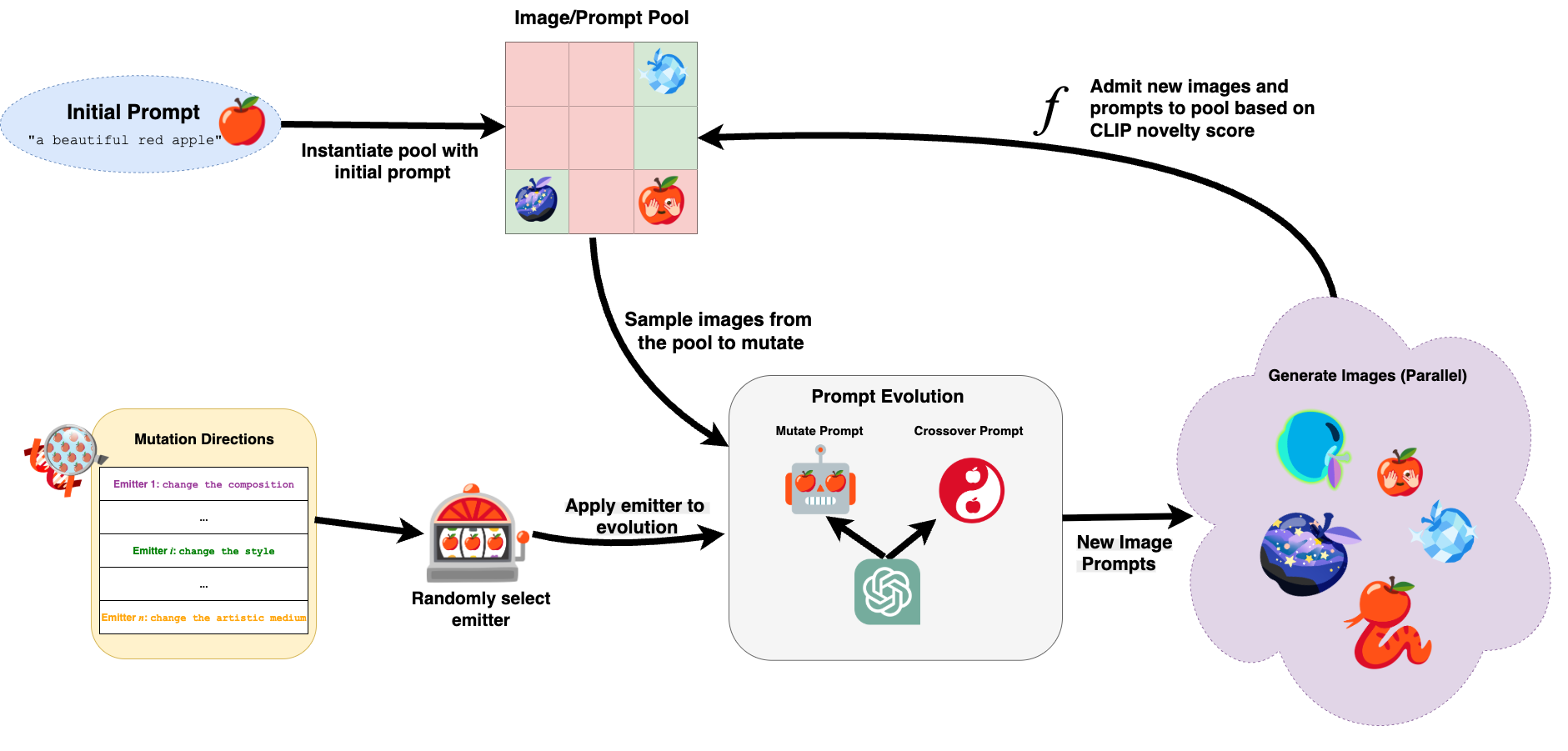}
  \caption{An overview of the \algname workflow.}
  \label{fig:flowchart}
\end{figure*}

\begin{table*}[b!]
\centering
\caption{Comparison of LLM evolution approaches \\ }
\begin{tabularx}{\textwidth}{lp{2.5cm}p{3cm}X}
\toprule
\textbf{Method} & \textbf{Initial Prompt} & \textbf{Objective} & \textbf{Evolutionary Approach} \\
\midrule
APE            & Multiple & Fitness & Crossover \\
EvoPrompt-GA      & Multiple & Fitness & Mutation, crossover \\
EvoPrompt-DE   & Multiple & Fitness & Mutation, crossover \\
PhaseEvo       & Multiple & Fitness & Directed mutation, crossover \\
QDAIF          & One      & Quality-Diversity & Directed mutation \\
Lluminate & One & Novelty & Mutation w/ creative strategies \\
\algname (ours)   & One      & Novelty & Directed mutation, crossover \\
\bottomrule

\end{tabularx}
\end{table*}

Inspired by related work, our approach evolves discrete, interpretable prompts through a mutation-selection loop. We specify a small number of simple emitters which significantly enhance image diversity. This enables transferability across downstream models regardless of their black-box nature or internal training objectives, providing a scalable, model-agnostic method for improving generation diversity.

%

The evolutionary cycle of \textsc{Wander} consists of three key components repeated over multiple generations: Emitter Selection, Prompt Evolution, and Pool Update. Each generation begins with a shared initial prompt population and evolves via LLM-generated transformations guided by emitters. Emitter effectiveness is continually reassessed based on diversity-driven feedback computed using image embeddings.


\subsection{Problem Formulation}
Given a pool $\mathbf{P}$ which contains up to $N$ prompt-image pairs $x_i = (p_i, I_i)$, we define a novelty score $f(x_i, \mathbf{{P}}) \in [0, 1]$ for each individual $x_i$ relative to the pool. The novelty score is typically a k-nearest neighbors mean embedding distance. Our algorithm's objective is then to generate new images and prune low-novelty images to produce a highly diverse final pool. We can frame this as maximizing the lowest novelty score in the pool,

$$
\mathbf{P}^* = \max_{\mathbf{P}} \left ( \min_i ( f ( x_i, \mathbf{P} ) ) \right ).
$$

\subsection{Proposed Method}

\textsc{wander} (Figure \ref{fig:flowchart}) begins with a pool of $n \leq N$ prompt-image pairs, and proceeds over $\mathcal{T}$ generations. In each generation, we perform a fixed number of mutations, each consisting of 3 steps; Emitter Selection, Prompt Evolution, and a Pool Update.

\paragraph{Initial Pool}
We instantiate the pool with $n \leq N$ copies of the input prompt. We then generate images for each prompt, creating $n$ starting individuals $x_1,\dots,x_n$.


\paragraph{Emitter Selection}
Emitters are predefined mutation strategies (e.g., ``change the composition'', ``adjust the lighting'', ``add elements'') which are included in the mutation prompt to direct evolution. A full list of emitters can be found in Table~\ref{tab:emitters}.





\begin{figure}[t!]
    \centering
    \includegraphics[width=1.0\linewidth]{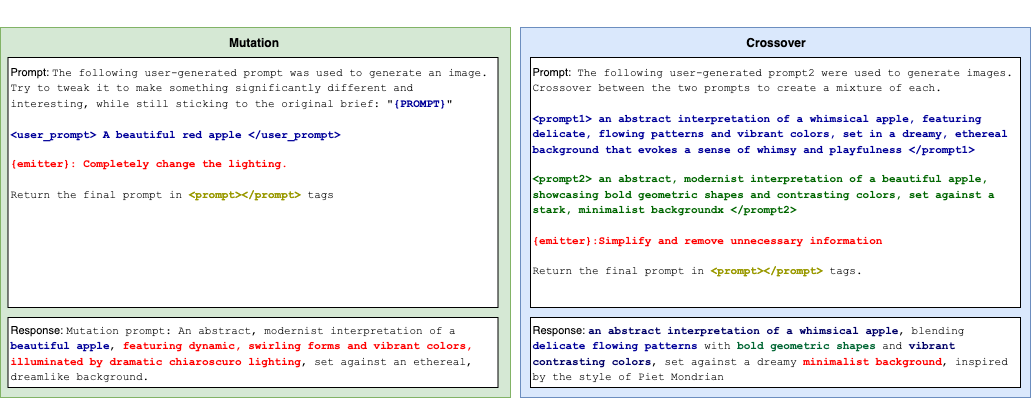}
    \caption{Examples of LLM prompt mutation and crossover}
    \label{fig:prompts}
\end{figure}

\paragraph{Prompt Evolution}
Once a mutation direction is selected, the framework applies one of two transformation techniques with a configurable probability (default 50\%): mutation or crossover \citep{guo2024connectinglargelanguagemodels, cui2024phaseevounifiedincontextprompt}. Directed by the chosen emitter, mutation modifies a single prompt, whereas crossover combines elements from two existing prompts to create a novel variation. This approach leverages large language models (LLMs) to generate high-quality variations, maintaining semantic coherence while fostering diversity. 
An example of this process is illustrated in Figure~\ref{fig:prompts}.

\paragraph{Pool Update}
For each newly evolved prompt $p_i$ we sample an image from the diffusion model to create candidate individuals $x_i$. For each image we then compute a CLIP embedding \citep{radford2021learningtransferablevisualmodels}. Following \citet{6793380}, we introduce an explicit novelty objective, measuring the average distance between an image embedding and its $k$-nearest neighbors in the pool. Formally, we define the novelty score for individual $x_i=(p_i,I_i)$ as:
\[
f(x_i, \textbf{P}) = \frac{1}{k} \sum_{j=1}^{k} d(I_i, I_{j}),
\]
where \( I_{j(i)} \) are the images of the \( k \) nearest neighbors of \( I_i \) in pool \( \textbf{P} \), and \( d(I_i, I_{j}) \) is the cosine distance. If the candidate has a higher novelty score than the current lowest in the pool, it then replaces the current lowest-scorer, ensuring that each generation progressively improves in diversity. 


This iterative refinement allows \textsc{Wander} to continuously explore and exploit the most effective mutation directions, leading to increasingly diverse and high-quality image generations.
\section{Experimental Setup}

\paragraph{Implementation Details}
We use GPT-4o-mini \citep{menick2024gpt4o} to perform prompt evolution, and FLUX-DEV \citep{flux2024} for image generation. For image and text embeddings we use OpenAI's CLIP-ViT-B-32 model \citep{radford2021learning}.

\paragraph{Baselines}
In our experiments, we compare \algname to several representative baselines in automatic prompt optimization, namely APE, EvoPrompt-GA, EvoPrompt-DE, PhaseEvo, QDAIF and Lluminate. For a comprehensive discussion of their underlying mechanisms and operational details, we refer readers to the Related Work section.

\paragraph{Tasks}
To compare \algname to other methods, we run all algorithms 10 times on the same prompt. To identify the impact of emitters, we conduct an ablation over emitter strategies to identify the impact of our design choices, which are run 10 times, for 10 prompts. Lastly, to compare random to bandit-driven selection, we run both methods for 30 generations to assess long-horizon performance.

We report LPIPS \citep{zhang2018perceptual}, Vendi score \citep{friedman2023the}, and a `Relevance' metric to evaluate image diversity and textual consistency. LPIPS is computed as the average pairwise perceptual distance between images based on deep feature representations. The Vendi score is defined as
\[
\mathrm{VS}(K) = \exp\left( -\sum_{i=1}^{n} \lambda_i \log \lambda_i \right),
\]
where \( \lambda_i \) are the eigenvalues of the normalized diversity matrix \( K/n \), constructed using pairwise cosine similarities between image embeddings. This score reflects the effective number of diverse samples in the pool. The Relevance metric is calculated as the average cosine distance between the text embeddings of the original and evolved prompts \citep{radford2021learningtransferablevisualmodels, hao2023optimizing, frans2021clipdrawexploringtexttodrawingsynthesis, tian2022modernevolutionstrategiescreativity}. For all three metrics, a higher score indicates better performance.

\begin{table*}[b!]
    \centering
    \caption{Results comparison of our method and existing baselines, 10 runs for each of 10 starting prompts for 100 total runs per method. The variants \algnamene\ and \textsc{-FE} refer to \algname\ with no emitters, and with a single fixed emitter per-run.}
    \label{tab:results}
\renewcommand{\arraystretch}{1}
\begin{tabularx}{\textwidth}{
    >{\raggedright\arraybackslash}p{2.6cm} 
    r@{\,${}\pm{}$\,}p{1.1cm}
    r@{\,${}\pm{}$\,}p{1.1cm}
    r@{\,${}\pm{}$\,}p{1.1cm}
    r@{\,${}\pm{}$\,}X
}
\toprule
\textbf{Method} 
& \multicolumn{2}{c}{\textbf{Vendi} $\uparrow$} 
& \multicolumn{2}{c}{\textbf{LPIPS} $\uparrow$} 
& \multicolumn{2}{c}{\textbf{Relevance} $\uparrow$} 
& \multicolumn{2}{c}{\textbf{Token Usage} $\downarrow$} \\
\midrule
EvoPrompt-DE & 1.42 & 0.04 & 0.51 & 0.01 & 0.292 & 0.001 & 38,243 & 4,514 \\
PhaseEvo     & 1.44 & 0.05 & 0.47 & 0.02 & 0.289 & 0.002 & 39,706 & 862 \\
APE          & 1.47 & 0.03 & 0.60 & 0.02 & 0.285 & 0.001 & 51,620 & 1,345 \\
EvoPrompt-GA & 1.49 & 0.02 & 0.56 & 0.01 & 0.295 & 0.001 & \textbf{1,828} & \textbf{19} \\
QDAIF        & 1.80 & 0.02 & 0.51 & 0.02 & \textbf{0.297} & \textbf{< 0.001} & 43,464 & 45 \\
Lluminate & 3.29 & 0.02 & 0.75 & 0.01 & 0.210 & 0.070 & 175,902 & 9,390 \\
\midrule
\algnamene   & 2.61 & 0.10 & 0.79 & 0.01 & 0.279 & 0.004 & 23,884 & 493 \\
\algnamefe   & 2.95 & 0.25 & 0.76 & 0.02 & 0.271 & 0.006 & 23,492 & 1,958 \\
\algname     & \textbf{3.60} & \textbf{0.09} & \textbf{0.80} & \textbf{0.01} & 0.272 & 0.003 & 24,347 & 649 \\
\bottomrule
\end{tabularx}
\end{table*}

\section{Results}


\begin{figure}[t!]
    \centering
    \begin{minipage}[t]{0.48\textwidth}
        \centering
        \includegraphics[width=\linewidth]{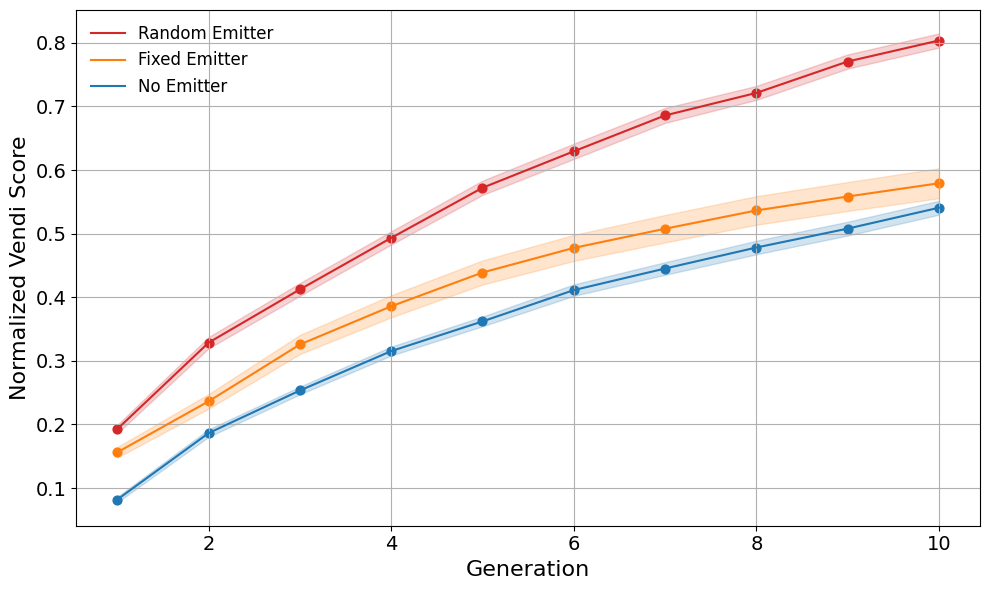}
        \caption{Ablation over emitter selection strategies. The results presented are averaged over 10 runs for each of 10 prompts (n=100 samples per method). For comparability, the Vendi score was min-max normalized per prompt.}
        \label{fig:ablation-emitter-strategy}
    \end{minipage}%
    \hfill
    \begin{minipage}[t]{0.46\textwidth}
        \centering
        \includegraphics[width=\linewidth]{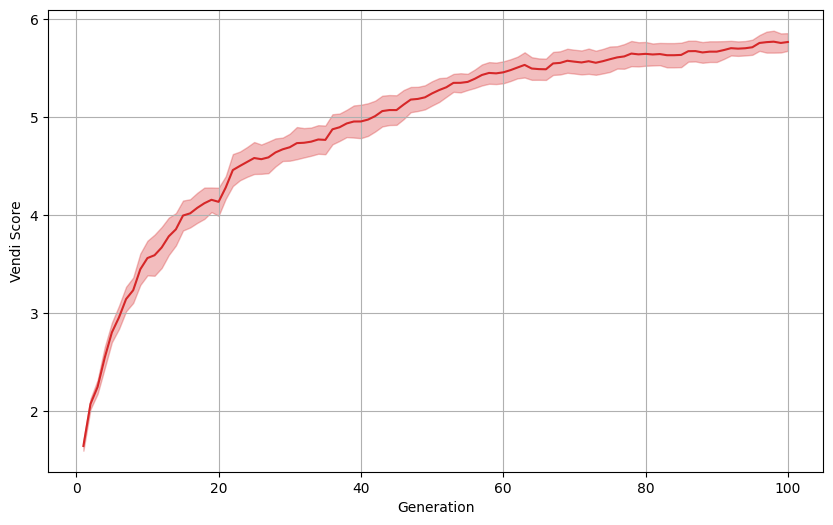}
        \caption{Over longer runs, the Vendi score consistently rises, plateauing around the $100^{\text{th}}$ generation. Averaged over 10 runs, shaded area indicates the standard error.}
        \label{fig:30gen}
    \end{minipage}
\end{figure}

\paragraph{\algname Achieves Superior Diversity and Efficiency.} 
Table~\ref{tab:results} presents a comprehensive comparison of our method with existing baseline approaches.  \algname achieves higher diversity than baselines with a Vendi score of $3.60 \pm 0.09$ and an LPIPS score of $0.80 \pm 0.01$, while requiring only $24,363 \pm 485$ tokens on average. In contrast, baseline methods such as Lluminate produced less diverse outputs (Vendi: $3.29 \pm 0.02$, LPIPS: $0.75 \pm 0.01$) while using significantly more tokens ($175,902 \pm 9,390$). Variants of \algname with no emitters ($2.61 \pm 0.10$ Vendi) and a single fixed emitter ($2.95 \pm 0.25$ Vendi) outperform QDAIF, demonstrating the efficacy of novelty search for diverse generation. Although the relevance score of \algname ($0.272 \pm 0.003$) is slightly lower than that of QDAIF ($0.297$), our qualitative analysis (examples in \autoref{sec:example-images}) indicates that generated images remain strongly aligned with the intended class, with rare exceptions discussed in \autoref{sec:limitations}. This suggests that the marginal reduction in relevance score does not compromise the practical usability of the final image pool.



\paragraph{Multiple Emitters Significantly Enhance Evolutionary Diversity.} In order to assess the impact of different emitter selection strategies, we conduct an ablation study involving a short evolutionary task spanning 10 generations. We evaluated several approaches, including a bandit-driven strategy, random selection, the use of a single fixed emitter per run, and a condition with no emitters. The results in Fig.~\ref{fig:ablation-emitter-strategy} indicate that employing multiple emitters leads to a substantial increase in the diversity of the final evolved pool compared to using a single, fixed emitter or no emitter at all. Image samples for different prompts are displayed in Appendix~\ref{sec:example-images}.

\begin{figure*}[h]
    \centering
    \includegraphics[width=1.0\linewidth]{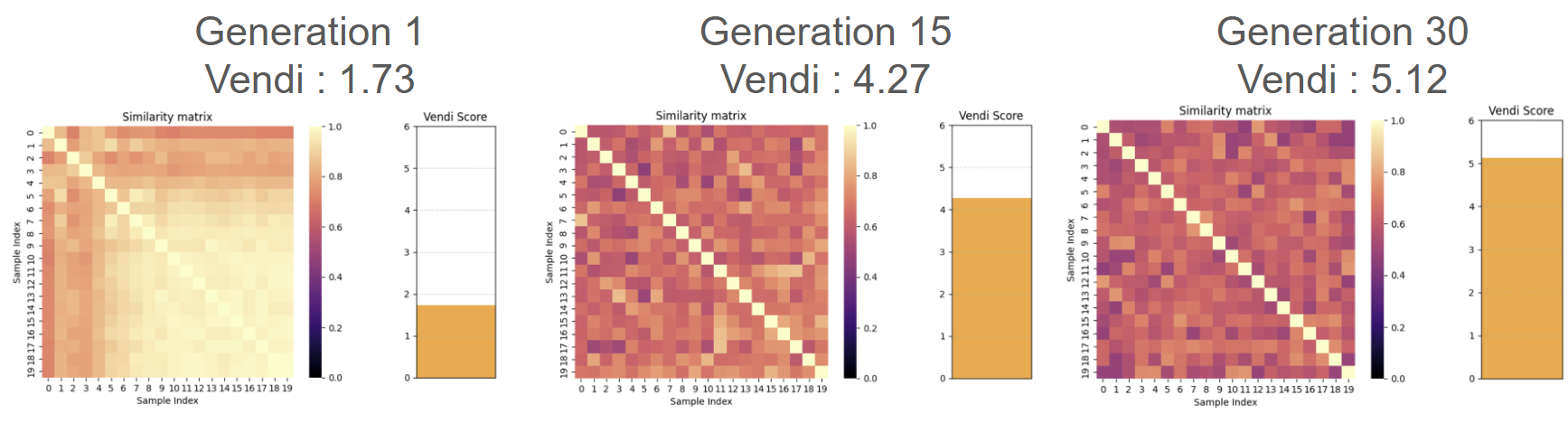}
    \caption{Similarity matrices of \algname image embeddings and Vendi scores at generations 1, 15, and 30.}
    \label{fig:sim}
\end{figure*}

\begin{figure}[h]
    \centering
    \includegraphics[width=0.5\linewidth]{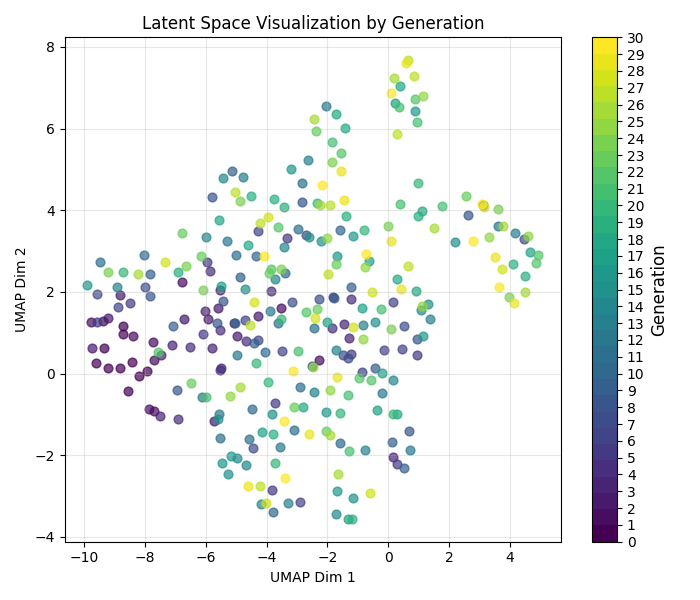}
    \caption{UMAP visualization of \algname image latents over 30 generations, each containing 10 generated images.}
    \label{fig:UMAP}
\end{figure}

\paragraph{Increased Diversity in Image Latent Space Through Evolution.} To understand the evolutionary dynamics within the image latent space, we visualized the image embeddings using Uniform Manifold Approximation and Projection (UMAP) \citep{mcinnes2020umapuniformmanifoldapproximation}. As shown in Fig.~\ref{fig:UMAP}, a clear trend of increasing diversity emerges across generations, demonstrating the impact of the evolutionary process on the latent space. Furthermore, the distinct spatial clustering of image samples from different generations in the UMAP visualization suggests a consistent evolution of their underlying latent representations. This observation is further supported by the similarity matrices and Vendi scores for generations 1, 15, and 30 in Fig.~\ref{fig:sim}. These results illustrate a decrease in pairwise image similarity and a corresponding increase in Vendi score as generations progress, quantitatively confirming the improved diversity of the generated image set over time.

\paragraph{More Capable LLMs Generate More Diverse Pools} For most results we use GPT-4o-mini as a cheap, fast prompt-mutation operator. To assess the impact of more capable LLMs for mutation, we compare different models from OpenAI over 10 runs, each of 20 generations. We find that more generally capable models are more effective prompt mutators, with OpenAI's o3 model achieving a 23\% higher Vendi Score than GPT-4o-mini. However, we also observe that reasoning models use around three times as many tokens as non-reasoning models. Results with the standard error are shown in Table \ref{tab:models}.

\begin{table}[ht]
\centering
\caption{Model comparison by Vendi score and token usage after 20 generations, averaged over 10 runs each, with the standard error. Token usage is higher than that shown in \autoref{tab:results} as \algname was run for 20 generations rather than the 10 used for the main comparisons.}
\label{tab:models}
\begin{tabular}{lcc}
\toprule
Model & Vendi Score $\uparrow$ & Token Usage $\downarrow$ \\
\midrule
GPT-4o-mini & $4.2 \pm 0.1$ & $\mathbf{61,402 \pm 1,309}$ \\
o4-mini     & $4.5 \pm 0.1$ & $227,655 \pm 7,114$ \\
GPT-4o      & $4.8 \pm 0.1$ & $78,067 \pm 2,031$ \\
o3          & $\mathbf{5.2 \pm 0.1}$ & $236,081 \pm 8,300$ \\
\bottomrule
\end{tabular}
\end{table}
\section{Limitations}
\label{sec:limitations}


\begin{itemize}
    \item \textbf{Relevance Drift:} The novelty objective can occasionally lead generated images to diverge from the initial prompt's core concept, roughly once per 5 runs, or 100 images. Mitigating this may require additional prompt tuning or an explicit relevance penalty during selection.
    \item \textbf{Human-Designed Emitters:} Emitters must be manually specified, which could bias or limit asymptotic diversity. The use of an LLM to generate emitters could uncover more effective mutation strategies without requiring any explicit human input beyond the initial prompt.
    \item \textbf{Aesthetic Evaluation:} Our evaluation focused on diversity (Vendi score and LPIPS) and prompt similarity. In early experiments, the Stable Diffusion 1.0 diffusion model was prone to generating low-quality images during evolution. However, as we did not observe this issue using FLUX-DEV, we did not assess it for generated images. However, evaluating the aesthetic quality of the generated images could provide a more complete picture of the efficacy of \algname.
\end{itemize}
\section{Conclusion}

This paper introduces \algname, a novel evolutionary framework designed to address the lack of diversity in text-to-image generation. Moving beyond simple aesthetic optimization, our method employs novelty search, using an LLM to mutate prompts guided by diverse emitters. Experiments confirm that CLIP embeddings serve as a useful component in novelty metric and that our bandit-driven emitter selection significantly enhances image diversity compared to baseline methods, particularly over extended runs. \algname provides an effective, adaptive strategy for generating varied image sets, supporting open-ended creative exploration with diffusion models.

\paragraph{Future Work} While this work demonstrates the effectiveness of \algname for image generation, several avenues remain for future research. This approach can be extended to any domains where meaningful distance metrics can be defined on latent representations, such as text and audio. For example, we used a text-based version of \algname to inspire the title of this paper (see \autoref{appendix:textgen}). We also hope to further investigate steerability of \algname; in this work we begin from simple prompts, but it may be desirable to constrain the direction of exploration more strongly. Finally, there are potential downstream applications which warrant further investigation, such as generating image model jailbreaks, or data augmentation for computer vision tasks.


\bibliography{acl2021}
\bibliographystyle{plainnat}

\appendix

\section{QDAIF Implementation}
\label{appendix:attempted}


In early QDAIF testing, we evaluated GPT-4o-mini and Qwen-2.5-VL as Vision-Language Models (VLMs) for image rating. The implementation includes a MAP-Elites grid defined by two image axes, for example detail and image style. To populate this archive, we prompted the mutation LLM to mutate a prompt towards a specified cell in the grid. We then prompted the VLM to evaluate generated images based on three criteria: quality, axis 1, and axis 2. Images were then assigned to specific cells within the grid according to the VLM's assessment. When multiple images were categorized into the same grid cell, the image with the highest quality score, as determined by the VLM, was retained to represent that cell. 

\begin{figure}[h!]
    \centering
    \includegraphics[width=0.5\linewidth]{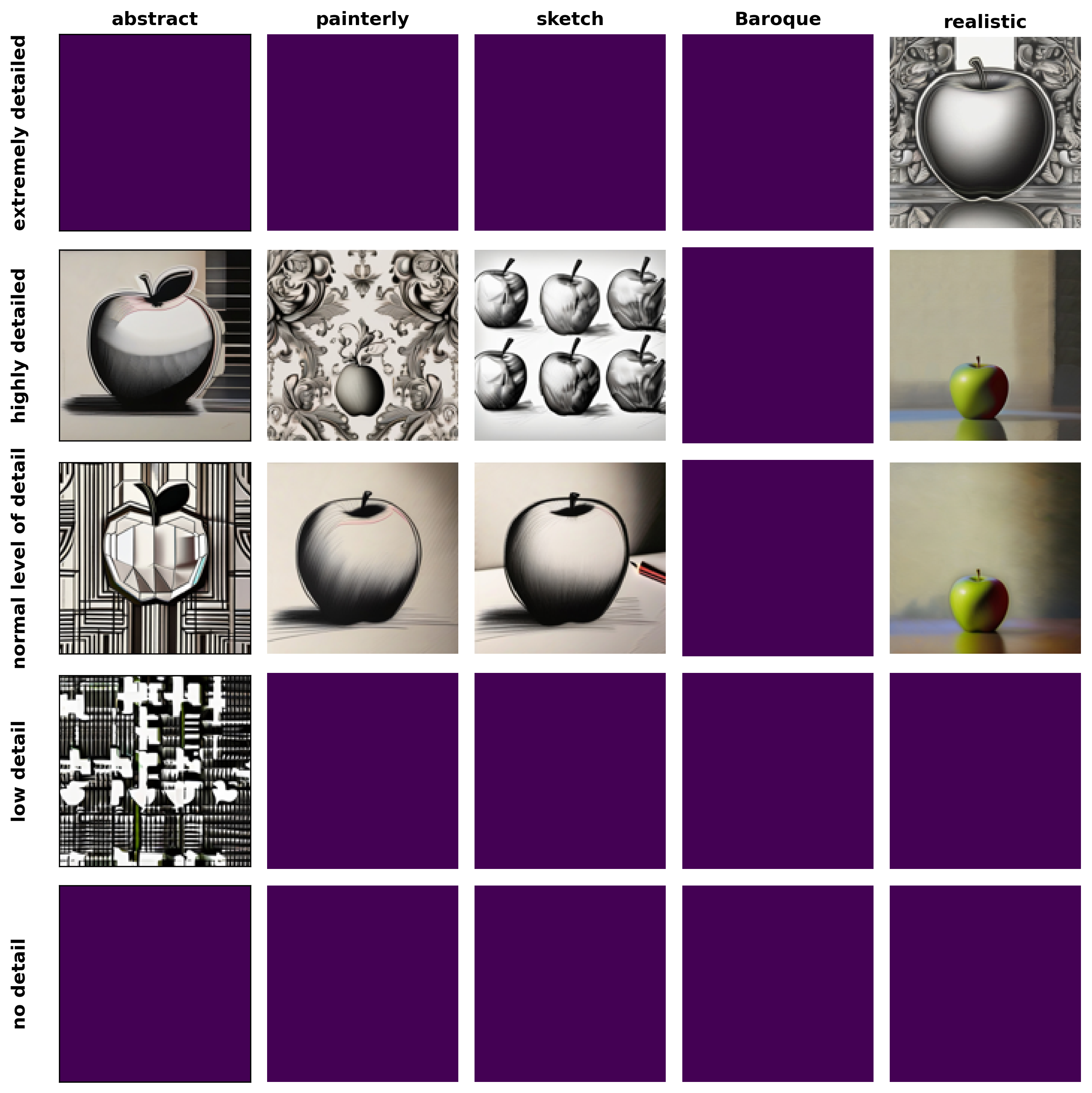}
    \caption{An example MAP-Elites grid after 20 generations of QDAIF using a VLM for feedback. The VLM allocates similar images to quite different quadrants, and gave aesthetic ratings for images inconsistent with qualitative evaluation.}
    \label{fig:qdaif-pool}
\end{figure}

Our findings indicated that the feedback provided by the VLM regarding image quality and MAP-Elites axes was not sufficiently nuanced or consistent to effectively guide the quality-diversity search. As illustrated in Fig.~\ref{fig:qdaif-pool}, some images are incorrectly categorized, or similar images are placed in different cells. These observations suggest that a key challenge we encountered stems from the inherent difficulties in using VLMs to effectively assess the complex characteristics of images for quality-diversity algorithms. Separately, we observed that defining suitable axes for such open-ended diversity tasks places an additional nontrivial requirement on a human user. For a comparison of QDAIF to our \algname approach, see Table \ref{tab:results}.

\newpage

\section{Example Images}
\label{sec:example-images}

\begin{figure*}[h]
    \centering
    \includegraphics[width=1\linewidth]{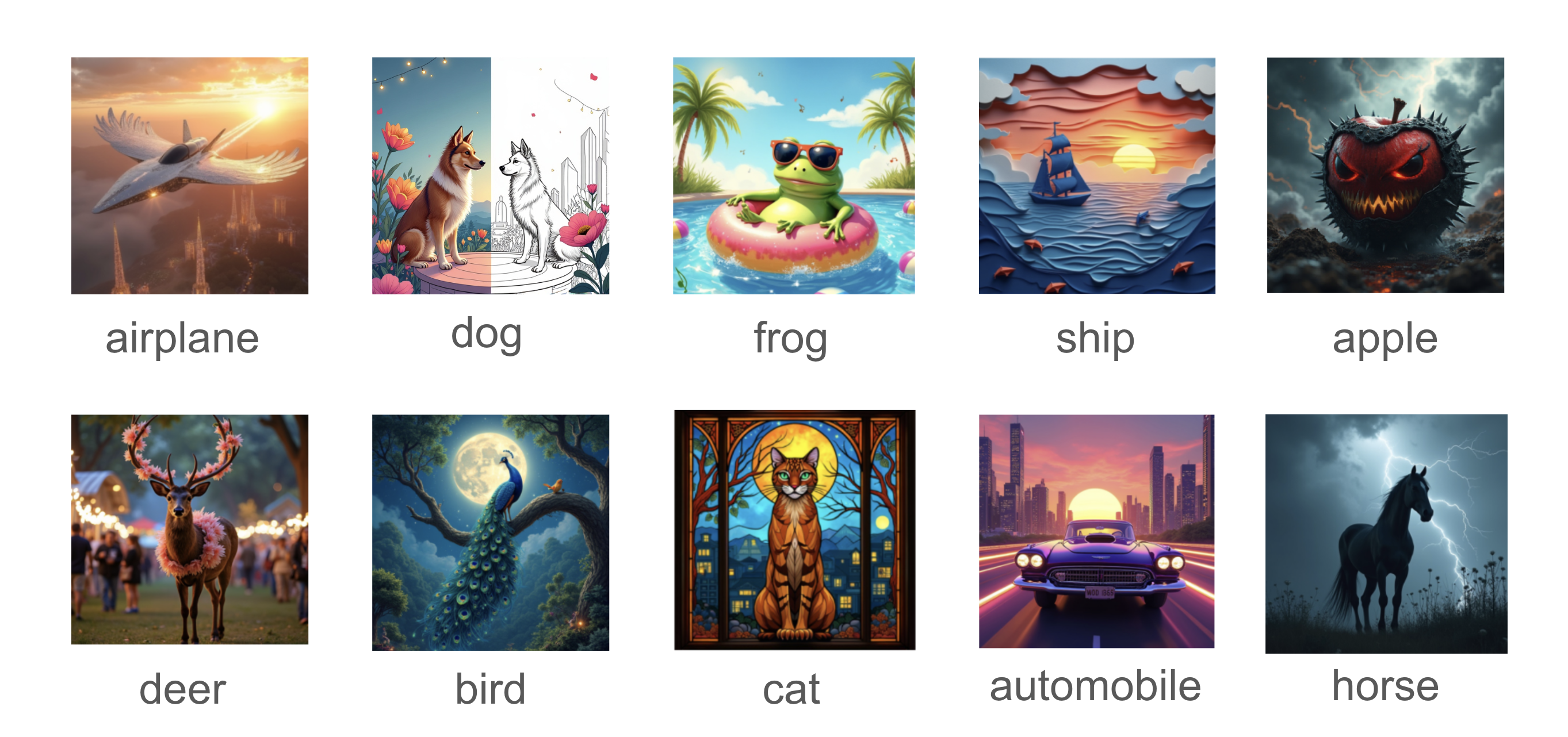}
    \caption{10 examples of novel images generated by \algname. Text indicates the input prompt.}
    \label{fig:fave-images}
\end{figure*}

\begin{figure*}[h]
    \centering
    \includegraphics[width=1\linewidth]{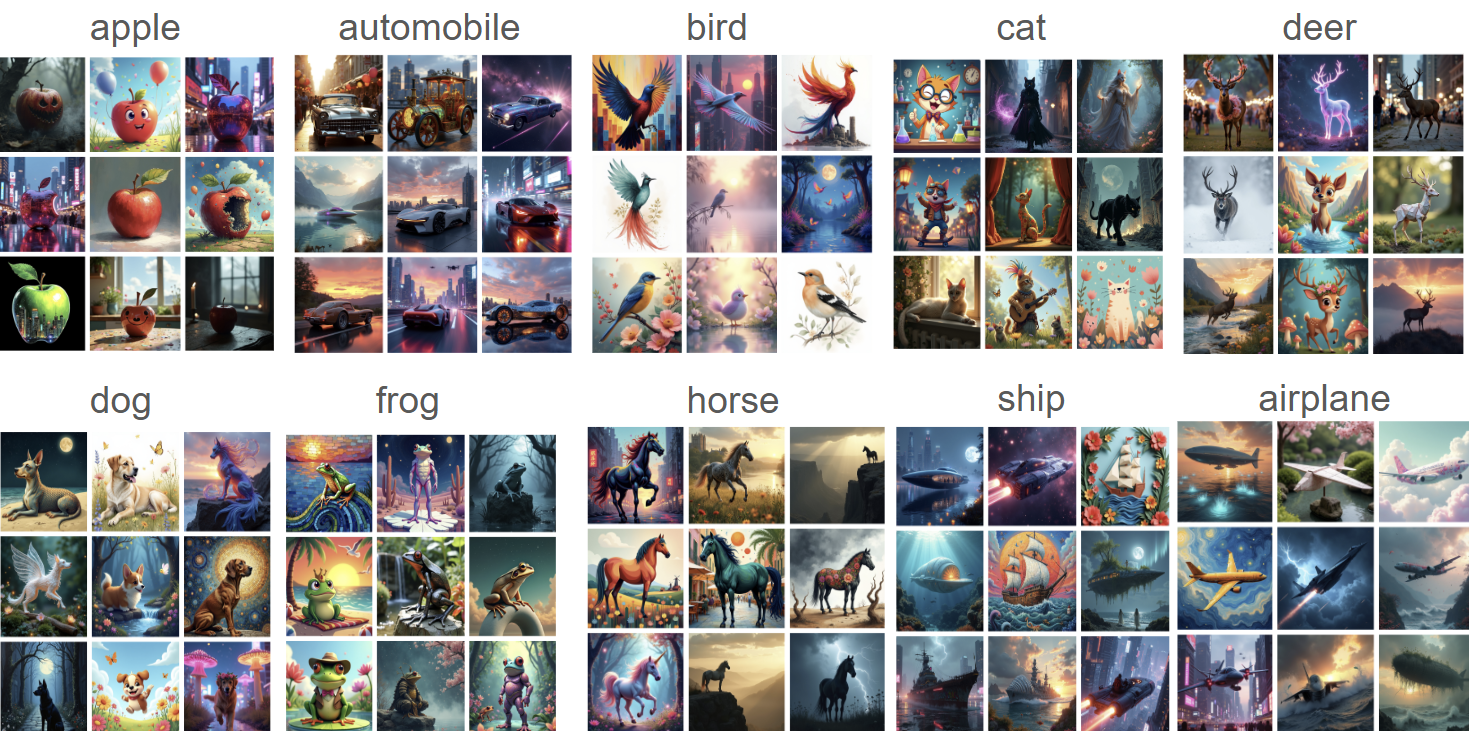}
    \caption{A selection of pools generated for different prompts}
    \label{fig:image-pool}
\end{figure*}

Figures ~\ref{fig:fave-images} and~\ref{fig:image-pool} show final pools of images generated by \algname. We use initial prompts inspired by CIFAR-10.



\newpage

\section{List of Emitters}
\label{appendix:emitters}
See Table \ref{tab:emitters} for a list of emitters used in all experiments.

\begin{table*}[h]
\centering
\caption{List of Emitters}
\begin{tabularx}{\textwidth}{lp{3cm}p{3cm}X}
\toprule
\textbf{Emitter} \\
\midrule
1. Completely change the composition. \\
2. Completely change the style. \\
3. Completely change the mood. \\
4. Completely change the lighting. \\
5. Completely change the atmosphere. \\
6. Completely change the artistic medium. \\
7. Add additional elements, while retaining the primary focus. \\
8. Simplify and remove unnecessary information. Be concise. \\
9. Come up with an artist to make it similar to. \\
10. Suggest a novel color scheme. \\
\bottomrule
\label{tab:emitters}
\end{tabularx}
\end{table*}

\begin{samepage}
\section{\algname for Text Generation}
\label{appendix:textgen}
We briefly experimented with the use of \algname for other modalities including text generation. We present results from a run creating ideas for paper titles for this work.

\textbf{Initial Prompt:} The title for an academic research paper which uses evolutionary algorithms with LLMs for creative text and image generation. Include only the title, no other information, less than 10 words.

\textbf{Text Results:}
\begin{itemize}
  \item ``Evolve \& Inspire: Algorithms Crafting Creative AI Journeys''
  \item ``Evolutionary Synergies: LLMs Ignite Creative Algorithmic Horizons''
  \item ``Algorithmic Brushstrokes: The Language Model Canvas''
  \item ``Imagining Pixels: Darwin Meets GPT in Artistry''
  \item Darwin's Brush: Evolutionary Artistry in AI
  \item ``Algorithmic Muse: LLMs Shaping Tomorrow's Artistic Landscapes''
  \item ``Evolving Algorithms: Creative Text and Image Synthesis''
  \item ``Artistry Evolution: LLMs Transforming Creative Frontiers''
  \item ``From Nature to Nurture: AI's Evolutionary Creative Leap''
  \item Enhanced Evolutionary Design: Text--Image Innovation with LLMs
  \item ``Crossing Codes: Merging AI Minds \& Genetic Blueprints''
  \item ``Revolutionizing Imagination: AI Marriages of Biology and Language''
  \item ``Pixels and Prose: Evolutionary Algorithms Meet Creative Machines''
  \item ``AI Artistry: Evolution's Algorithms in Creative Harmony''
  \item ``Evolutionary Synergy: LLMs in Creative Text--Image Generation''
  \item LLMs Unleashed: Evolving Art through Algorithms
  \item ``Evolving Creativity: Harnessing LLMs for Dynamic Content Generation''
  \item Evolutionary AI: Harmonizing Nature and Imagination
  \item ``Evolutionary LLMs: Innovative Creative Geneation with Algorithms''
  \item ``Co-Evolutionary Creativity: LLMs Transforming Media''
\end{itemize}
\end{samepage}





\end{document}